# Approaches to Corpus Creation for Low-Resource Language Technology: the Case of Southern Kurdish and Laki

**Sina Ahmadi**♠  **Zahra Azin**♣  **Sara Belelli**♦  **Antonios Anastasopoulos**♠

♠Department of Computer Science, George Mason University, Fairfax, VA, USA
♣School of Linguistics and Language Studies, Carleton University, Canada
♦Università degli Studi della Tuscia, Viterbo, Italy

{sahmad46,antonis}@gmu.edu, taraazin@cmail.carleton.ca, sarabelelli@gmail.com

## Abstract

One of the major challenges that under-represented and endangered language communities face in language technology is the lack or paucity of language data. This is also the case of the Southern varieties of the Kurdish and Laki languages for which very limited resources are available with insubstantial progress in tools. To tackle this, we provide a few approaches that rely on the content of local news websites, a local radio station that broadcasts content in Southern Kurdish and fieldwork for Laki. In this paper, we describe some of the challenges of such under-represented languages, particularly in writing and standardization, and also, in retrieving sources of data and retro-digitizing handwritten content to create a corpus for Southern Kurdish and Laki. In addition, we study the task of language identification in light of the other variants of Kurdish and Zaza-Gorani languages.[1]

## 1 Introduction

Language and linguistic data play a critical role in documenting and preserving endangered and under-represented languages. Indispensable to computational methods in language technology, data also enables the development of tools and applications, such as speech recognition and machine translation, that can support the revitalization and promote the usage of such languages. As such, speakers of endangered and under-represented languages ultimately have the opportunity to share their language and cultural heritage with future generations. Despite the fascinating advances in natural language processing (NLP) in recent years, particularly in working with very limited data in low-resource setups (Hedderich et al., 2021), collecting data for endangered and less-resourced languages remains a challenging task.

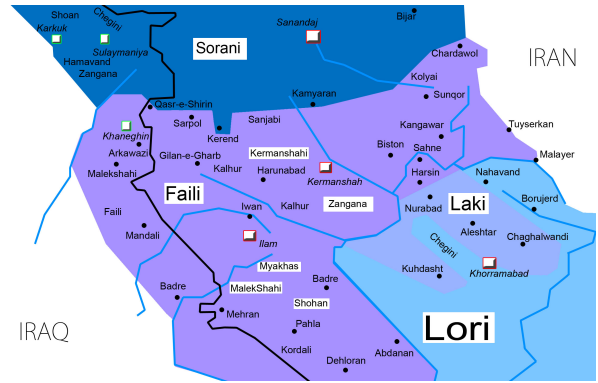

Figure 1: Territories where Central Kurdish (dark blue), Southern Kurdish (violet), Laki (pale violet) and Lori (blue) are mainly spoken. Based on Fattah (2000)

In this paper, we focus on Southern Kurdish (sdh in ISO 639-3) which is one of the main varieties of Kurdish spoken by an estimated 3.7 million speakers in the provinces of Kirmaşan, also spelled as Kermanshah, and Ilam in Iran, and across the adjoining border regions of Iraq (Eberhard et al., 2022). We also shed light on Laki (lki) spoken by a few hundred thousand speakers in the same regions (Aliakbari et al., 2015). Kurdish in general and Southern Kurdish and Laki in particular, have faced various discriminatory language policies that have led to pernicious sociolinguistic effects on language attitudes and heritage language maintenance such as the lack of children's proficiency in Southern Kurdish and limited usage of the language in writing (Sheyholislami, 2012; Tamleh et al., 2022).

As such, Southern Kurdish speakers have been facing centuries-long pressures of Persian as the only official language of Iran and the administratively dominant one, which led to various phenomena of language shift and change (Sharifi et al., 2013; Weisi, 2021; Yarahmadi, 2022). Although other varieties of Kurdish have not been immune from such policies, their larger population and strong Kurdish national and political identity be-

---

ing supported for a long time have been beneficial to promote the language, especially for the adjacent Central Kurdish speakers, also known as Sorani (`ckb`), in Iran and Iraq (Sheyholislami, 2010). Nevertheless, studies show that there is a positive attitude towards the Kurdish language and identity in communities of Southern Kurdish speakers as well (Rezaei and Bahrami, 2019; Sheyholislami and Sharifi, 2016).

In stark contrast to other varieties of Kurdish, particularly Northern Kurdish (`kmr`), also known as Kurmanji, and Central Kurdish, Southern Kurdish varieties and Laki have not received much attention in linguistics or computational linguistics. Moreover, for Southern Kurdish and Laki, there are relatively much fewer digital resources available, not to say near zero, and both face practical challenges in writing. On the other hand, studying these varieties *in loco*, i.e. linguistic fieldwork, is not always a viable solution given the geopolitical conditions of the region and limitations in cost and time.

**Contributions** This paper sheds light on creating a corpus for Southern Kurdish and Laki. We discuss possible approaches that can be taken to tackle corpus creation for under-represented and endangered languages by relying on local content creation media and also, fieldwork. Our corpus contains over 2 million tokens in Southern Kurdish and Laki. In addition to an intrinsic evaluation of the corpus, we also analyze the corpus in a qualitative way and extend our analysis to the task of language identification.

## 2 Southern Kurdish vs. Laki

### 2.1 Language Classification

Kurdish identity has been shaped by centuries of history and a strong attachment to land and culture. However, the quest for defining the Kurdish language has been a complex and challenging process, shaped by political and social factors.

Although it is difficult to define Kurdish as a homogeneous language, and it is debatable whether it should be described as a continuum of dialects, or rather as a *Sprachbund* (Jugel et al., 2014), there is broad consensus on the fact that Northern Kurdish, Central Kurdish and Southern Kurdish are the three main varieties of Kurdish as described by McCarus (2009), Edmonds (2013) and many others.

Other Iranic languages of Kurdistan, such as Zazaki (`zza`) and Gorani (`hac`) are commonly considered distinct from Kurdish even though their speakers share close cultural bonds with neighboring Kurdish communities and not rarely consider themselves as ethnically Kurds (Haig and Öpengin, 2014, cf.). That said, these two are sometimes referred to as the two other dialects of Kurdish (Eppler and Benedikt, 2017). Moreover, the classification of Laki as the southernmost variety of the Kurdish language cluster is a debated issue. On the other hand, there is full scholarly consensus on the fact that Luri (also spelled Lori, `lrc`/`luz`) is a Southwestern Iranic language, despite the common misconception of it being a variety of Kurdish (Anonby, 2004). Nevertheless, Lori and even more so Laki might show convergence phenomena with neighboring Southern Kurdish dialects and vice versa

In this paper, we focus on the varieties of Southern Kurdish that are spoken in the province of Kermanshah to which we refer as Kermanshahi (also spelled Kirmaşanî) and those that are spoken in Iraq. Southern Kurdish is described in the literature as a diverse group of Kurdish parlances that can be clustered into several dialect groups, among which Garrusi, Kordali, Kalhori, and Feyli as outlined by Belelli (2019, 2021). It is worth noting that here we use the term 'Feyli' as a collective denomination for some Southern Kurdish dialects spoken in border regions of Iraq and the capital Baghdad, although we recognize that the use of the term as a language label has problematic sides which cannot be further discussed here. Similarly, we also take into consideration so-called Laki-Kermanshahi varieties, which were considered as part of Southern Kurdish in (Fattah, 2000) but are perhaps better described as mixed varieties intermediate between Southern Kurdish and the Laki of northern Lorestan and eastern Ilam.

### 2.2 Morphosyntactic Comparison

On the differences between Northern and Central Kurdish varieties, many studies have been conducted (Matras, 2019; Esmaili and Salavati, 2013, cf.). Similarly, Belelli (2021, p. 17) lays out the major differences between Southern Kurdish and other Kurdish varieties. However, the differences between Southern Kurdish and Laki are less discussed in the literature.

Although Southern Kurdish shows morphological similarities with both Northern Kurdish and Central Kurdish, it is closer to the latter, not having morphological markers of gender and case. Moreover, Southern Kurdish is unique within Kurdish varieties, not showing forms of tense-sensitive alignment, unlike the ergative properties of Northern and Central Kurdish. On the other hand, the differences between Southern Kurdish and Laki are less discussed in the literature, although Laki-Kermanshahi parlances have been observed to form a continuum in which the number of Laki-like features adds up proceeding from cities of Sahne towards Harsin, or rather Southern Kurdish-like traits progressively increase proceeding in the opposite direction (see Figure 1). The dialect of Harsin shows the highest level of morphological and lexical similarity with Laki "proper", while that of Sahne is the closest to Southern Kurdish.

| Part-of-speech | | Northern Kurdish | Central Kurdish | Southern Kurdish | Laki |
|---|---|---|---|---|---|
| **Noun** | **DEF** M | ∅ | | -ege, -eke | -e, -ke |
| | F | ∅ | -eke | | |
| | PL | ∅ | -ekan | -egan, -ekan, -eğan, -eyle(ge) | -ele |
| | **INDF** M | -ek | | -î, -îg, -îk, -îğ | -ê, -î, -îk |
| | F | -ek | -êk | | |
| | PL | -in | -an , -gel | -eyl, -gel, -ğel, -an | -el |
| **Verb** | **INF** | -in | -in | -in | -in |
| | **PROG** | di- | e- , de- | ∅, di-, e- | (-e) me- |
| | **SBJV** | bi- | bi- | bi- | bi- |
| **Adjective** | **NEG** | ne-, na- | ne-, na-, me- | nye-, ne-, na-, nî- | nime-, ne-, nî- |
| | **COMP** | -tir | -tir | -tir, -tirek, -tirig | -tir |
| | **SUP** | -tirîn | -tirîn | -tirîn | -tirîn |

Table 1: A comparison of affixes in varieties of Kurdish and Laki. Abbreviations are according to Leipzig Glossing Rules (Comrie et al., 2008). For consistency, the Kurdified Latin script of Bedirxan is used for all. Nominal affixes are merged for variants lacking grammatical gender.

Regarding Laki, among the typical Laki-like traits of Harsini and other Laki-Kermanshahi dialects, such as Payrawandi, Sahne'i, are the presence of phonemic /v/ as in *vitin*[2] vs Southern Kurdish *witin* 'to say', the presence of phonemic /δ/ as in *döm* 'tail' vs. Southern Kurdish *dom, dim* and variants, the form *homa* of the second person plural pronoun, a discontinuous indicative marker =*a ma-* (except Sahne having *a-* as some Southern Kurdish dialects), the use of (post)verbal particles instead of common Kurdish preverbs, such as *ör* instead of *hal* 'up', the use of different adpositional forms, such as *va* 'to, at' vs. Southern Kurdish *wa* 'to', *la/da* 'at' and the reflexive marker *wiž* as opposed to Southern Kurdish *xwa* and variants.

On the other hand, Harsini and the rest of Laki-

Kermanshahi dialects have a form of the second person singular and plural verbal endings -*î(t)/-îtin* which differs from Laki -*î(n)/-înān*, -*înō(n)* (and of isomorphic clitic copula forms), and a form of the third person plural clitic pronoun =*yān* differing from Laki =*ān*, =*ō(n)*. Moreover, all Laki-Kermanshahi dialects share with Southern Kurdish the absence of forms of agentiality in the conjugation of past transitive verbs, which is otherwise a distinctive feature of Laki, as well as of Central Kurdish. Table 1 summarizes some of the frequent affixes.

### 2.3 Lexical Differences

Concerning Southern Kurdish and Laki, there are a series of words that are distinctive to Laki, among which *āyl* 'child', *pit* 'nose', *lam* 'stomach', *sīr* 'sated', *gojar* 'small' vs. Southern Kurdish *minaɫ*, *lūt, zik, tīr, büčik/g*, respectively (Aliyari Babolghani, 2019). It must be noted that due to the sociolinguistic and geopolitical conditions, Southern Kurdish and Laki, as virtually all other regional languages, have been historically sensible to lexical borrowing from dominant languages, especially Persian and Arabic.

### 2.4 Writing

Although the two main scripts currently used for writing Kurdish, that is the Latin-based '*Hawar*' or '*Bedirxan*' script and the Perso-Arabic script of Central Kurdish are also adapted for writing in Southern Kurdish, with distinct graphemes such as <ۊ> (U+06CA), these scripts are not widely used among speakers who rely on a the administratively-dominant language's writing system in practice, i.e. that of Persian or Arabic (Ahmadi et al., 2019; Filippone et al., 2022). In the same vein, Laki lacks a standard script or orthography.

Consequently, this adds to the complexity of the situation in which, a collected corpus should be written in a customized way or based on the script of a closely-related language, in this case, Central Kurdish.

## 3 Related Work

Although a less-resourced language, Kurdish has increasingly received attention in the past few years in language technology with tools such as the Kurdish language processing toolkit (Ahmadi, 2020b), services such as Google Translate[3] and models and benchmarks in NLP such as the FLORES-101 (Goyal et al., 2022) and NLLB (Costa-jussà et al., 2022). However, these solely include Northern and Central Kurdish but neither Southern varieties nor Laki.

Similarly, Wikipedia as an important resource to document languages is only available for Northern and Central Kurdish while Southern Kurdish and Laki along with other adjacent under-represented languages Gorani and Luri are not supported yet. Ahmadi et al. (2019) study the available lexicographical resources for Kurdish varieties and, as illustrated in Figure 2, find that among the 71 dictionaries and terminological resources available for Kurdish, Laki and Zaza-Gorani languages in electronic and printed forms, only 13.6% have content for a Southern Kurdish variety or Laki.

Previously, some linguistic aspects of Southern Kurdish have been studied such as phonology (Kord Zafaranlu Kambuziya and Sobati, 2014), typology (Dabir-Moghaddam, 2012), morphology (Belelli, 2022) and dialectology (Fattah, 2000). Belelli (2021) studies Laki and describes its complex relationship with Southern Kurdish and also documents a Laki variety by collecting a lexicon and a corpus through fieldworking.

Considering resources for language technology, Azin and Ahmadi (2021) create an electronic dictionary in Ontolex-Lemon containing 14,326 entries of varieties of Southern Kurdish in addition to Laki and Luri languages. In this resource, entries are represented in both scripts commonly used for Kurdish, even though the Latin orthography is not much used for Southern Kurdish, in addition to

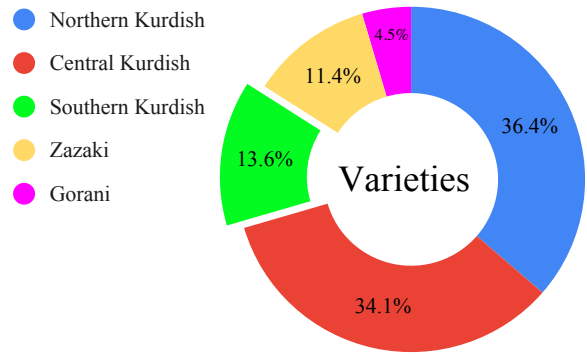

● Northern Kurdish
● Central Kurdish
● Southern Kurdish
● Zazaki
● Gorani

Figure 2: Percentage of the existing lexicographical resources for Kurdish varieties among which only 13.6% (<10 references) focus on Southern Kurdish and Lori.

translations in Persian and Central Kurdish. To the best of our knowledge, this dictionary is the only electronic resource for Southern Kurdish of considerable size. Similarly, Amani et al. (2021) collect audio samples for Kurdish spoken dialect recognition using radio and television contents among which 11 hours are collected for Southern Kurdish. Table 2 summarizes Kurdish and closely associated languages, along with some of the major corpora that have been previously created for them.

In addition to language documentation, corpora are crucial resources in many other applications such as language learning and teaching (Tribble, 2015), machine translation and syntactic parsing. To the best of our knowledge, no corpus of considerable size has been created for Southern Kurdish and Laki that is written in any of the two conventionalized Kurdish scripts.

## 4 Methodology

To remedy the lack of data for Southern Kurdish and Laki, we follow three approaches that are described in this section.

### 4.1 Radio Shows

In presence of local media, we resort to a local radio broadcaster in Kermanshah province (Iran) which is in majority inhabited by native speakers of Southern Kurdish. Upon our request, we could collect a set of handwritten scenarios of radio shows in Kermanshahi varieties of Southern Kurdish. The scenarios cover educational, cultural and daily topics and primarily target audiences in rural areas. Therefore, a rich native vocabulary of Southern Kurdish is employed with very few instances (if any) of code-switching to Persian or ex-tensive lexical borrowing from Persian.

---

[3] https://translate.google.com

| Language | 639-3 | Wikipedia | Common Scripts | Corpora |
|---|---|---|---|---|
| Northern Kurdish (Kurmanji) | kmr | ku | Latin, Central Kurdish | (Esmaili and Salavati, 2013; Ataman, 2018; Matras, 2019) |
| Central Kurdish (Sorani) | ckb | ckb | Central Kurdish, Latin | (Esmaili et al., 2013; Abdulrahman et al., 2019; Veisi et al., 2020; Ahmadi et al., 2020; Matras, 2019) |
| Southern Kurdish | sdh | – | Central Kurdish, Persian | (Fattah, 2000) |
| Gorani | hac | – | Central Kurdish | (Ahmadi, 2020a) |
| Zazaki | zza | diq | Latin | (Ahmadi, 2020a) |

Table 2: Description of Kurdish varieties along with Zazaki and Gorani with some of the existing corpora and scripts ordered based on popularity. Central Kurdish script refers to the Kurdified Perso-Arabic script commonly in use in Central Kurdish.

The radio shows dataset consists of 18 scenarios written for a local radio station in the city of Kermanshah. The scenarios are written for talk shows and short comedies and broadcasted from the same radio channel. The scenarios are written for 10 to 15-minute-long programs. Most of the programs are written in the form of dialogues which makes the dataset a good fit for future discourse analysis studies.

The original scenarios were written by hand, using Persian script and orthography. We asked three Southern Kurdish speakers to type the scenarios using the Central Kurdish Perso-Arabic script. This enables us to compare the data with materials written in other varieties of Kurdish. The manually typed data were then reviewed for possible inconsistencies in the writing form used by the typists.

## 4.2 News Articles

In our second approach, we follow the approach of Ahmadi (2020a) to crawl content from a news website to document Southern Kurdish varieties spoken in Iraq. We found a local news website[4] that publishes news articles in a few languages including Feyli. Overall, 15,985 articles are crawled in HTML and converted to text. Following this, we carry out text preprocessing by unifying character encoding using regular expressions, cleaning the raw text by removing private information such as email addresses and text formatting and categorizing the raw text based on the topic of the article, mainly in culture, politics and Kurdistan categories.

As metadata, we provide the source, topic, title and date for the collected articles.

---

[4]https://shafaq.com

## 4.3 Fieldwork

Finally, we rely on fieldwork to document Laki and create a corpus of oral texts in the language variety spoken in Harsin city in Kermanshah province in Western Iran, belonging to the so-called Laki-Kermanshahi (or Laki-Kirmashani) dialect cluster, identified as intermediate between Southern Kurdish and Lorestani Laki (Belelli, 2021). The content of the Harsini textual corpus is typologically uniform and includes seven traditional narratives – five folktales and two anecdotes – in the form of monologues, recorded from four speakers (three female and one male) native to Harsin or the neighboring village of Parive. The texts are manually transcribed following a conventional transcription system based on the tradition of Iranian linguistics, divided into numbered annotation units, and translated into English. One of the seven texts is further interlinearized with morpheme-by-morpheme glosses.

As there is no standard writing system or orthography for Laki, using Persian script or the Kurdified scripts for Laki remains optional rather than conventional choices. This said, transliterating the corpus is possible given the consistency in the phonetic transcription.

## 5 Results and Analysis

In this section, we carry out an intrinsic evaluation of our corpus alongside presenting a qualitative analysis. We also extend our analysis to the task of language identification.

## 5.1 Quantitative Analysis

The collected data contains 16,003 documents written in varieties of Southern Kurdish and seven narratives in Laki-Kermanshahi. Table 3 presents the number of articles, tokens, types, and type characters in our collected corpus. To calculate types, i.e. unique tokens, we exclude punctuation marks, digits, and sentences tentatively flagged as code-switching, such as religious quotations in Arabic or poems in Persian. Additionally, we use regular expressions for tokenization.

Since the vocabularies of the selected varieties have much in common, we also calculate the average type length as an indicator of the morphological complexity of word forms. Although the smaller size of Kermanshahi and Laki data might not reveal much about the morphological intricacies of these varieties, the average word lengths of 6.57 of Kermanshahi and 6.45 of Laki seem to be at odds with an average length of 8.8 of types in Feyli. In comparison to the other varieties of Kurdish and Zaza-Gorani languages, Southern Kurdish appears to have longer word forms with an average length overall. According to Ahmadi (2020a), Northern and Central Kurdish have an average length of 4.8 and 5.6, respectively. Similarly, Zazaki and Gorani have an average length of 4.84 and 5.50.

We think that this remarkable difference in word length can be due to a) the orthography of the Southern Kurdish corpus, texts in Feyli in particular, b) conventions in writing multiword expressions as in بیسەروشوونکیریاگ (*bîserûşûnkiryag*) 'doomed' composed as بی-سەر-و-شوون-کریاگ and c) excessive concatenation of words as in گورانیچرلوبنانی گورانیچر لوبنانی (*goranîçirr Lubnanî*) 'Lebanese singer'. We also notice that conjunction و (*û*) 'and' and prepositions like لە (*le*) 'in' are sometimes merged with the preceding or succeeding word, as in پەلامارەدەریلمو instead of پەلامارەدەریلە و (*pelamardereyle û*) 'attackers and' or لەشقامێگ instead of لە شقامێگ (*le şeqamêk*) 'in a street'. More importantly, affixes in Southern Kurdish, as shown in Table 1, are longer than the ones in Northern and Central Kurdish resulting in a higher average word length.

Additionally, we calculate the rank-size distribution in Pewan corpus for Northern and Central Kurdish (Esmaili et al., 2013) and Zaza-Gorani corpus (Ahmadi, 2020a) along with our Southern Kurdish data (merged Kermanshahi and Feyli). According to Zipf's Law (Zipf, 1999), in such a distribution "the length of a word tends to bear an inverse relationship to its relative frequency." This is illustrated in Figure 3 where the curves for each corpus start with the most frequent words (seen as dots), then words with a rank of 10 to 10000 smoothly diminish in frequency and finally, the majority of words appear at the bottom segment with lower frequencies (<10). We could not include the Laki data since this distribution requires a relatively big corpus to be valid.

| Number (#) | Kermanshahi | Feyli | Laki |
|---|---|---|---|
| articles | 18 | 15,985 | 7 |
| tokens | 10,127 | 2,182M | 6,340 |
| types | 3,248 | 179,208 | 2,074 |
| characters | 21,359 | 1,591M | 13,378 |
| average length | 6.57 | 8.8 | 6.45 |

Table 3: Statistics of the collected data based on varieties of Southern Kurdish (M refers to million). The number of characters and the average length are calculated based on the types.

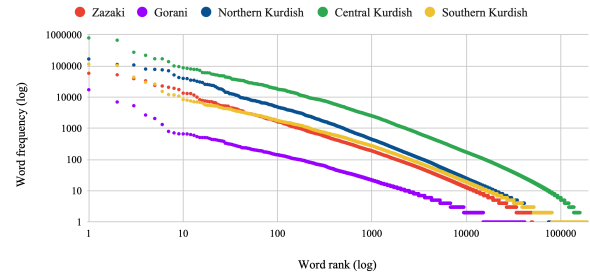

Figure 3: Zipfian distribution of Pewan corpora of Northern and Central Kurdish, a corpus of Zaza-Gorani and our corpus of Southern Kurdish (Kermanshahi and Feyli merged).

## 5.2 Qualitative Analysis

From a qualitative point of view, since the collected data fall into distinct textual genres, one would obviously expect differences in textual structure, lexical choices, and the level of formality. However, in the case of Southern Kurdish varieties spoken in Iraq and Iran, the most notable differences are related to the use of words classifiable as borrowings from other regional languages. Interestingly, varieties of Southern Kurdish used in Iraq tend to rely on the vocabulary of Central Kurdish as well, particularly when it comes to the terminology, while those in Iran rely more on the Persian vocabulary.

Based on the textual structure of the news articles collected for Feyli, each article has a headline that represents a concise version of the content using active voice. A collection of the headlines provides an exceptional resource for discourse analy-sis and further linguistic studies of the corpora. On the other hand, the data collected for Kermanshahi contains both monologues and dialogues written for radio shows. The dialogues that are written for comedy shows are in the form of two-way conversations between voice actors about social and cultural issues. The informal language used in the shows simulates real-world interactions between two speakers of Kermanshahi providing an opportunity for future conversation analysis. In the same vein, the narrative in the Laki data provides information useful to analyze folkloric stories.

Furthermore, Zipf's Law also states that the most frequent words in a language are the shorter ones due to economic factors (Sigurd et al., 2004). Table 4 provides the most frequent words in the selected corpora. Although many words among the most frequent ones have less than three characters, such as *û* 'and', *li/le/ce/de* 'in' and *bo/ara/ařā* 'for', many other words like *Kurdistan* and *Iraq* appear frequently indicating the topics of the texts and also, the bias and a lack of diversity in domains. This is also affected by the orthography of the language as postpositions like *ra* in Zazaki and *de* in Northern Kurdish appear frequently, while the equivalent ones in Central Kurdish as *da* and *ewe* don't appear so. Nevertheless, the most frequent words in Laki data show elements from the narratives such as *muše* 'IND-SAY.PRS-3SG'.

Despite sharing many linguistic features, the variations between Southern Kurdish varieties and Laki-Kermanshahi are not negligible. As previously discussed, the scarcity of language data and a lack of a writing system for this branch of Kurdish are among the reasons we still do not have a clear picture of the extant variation in the written forms of its sub-varieties.

### 5.3 Language Identification

Language identification or detection is the task of detecting the language in which a sentence is written. This task is used in many downstream applications in NLP such as sentiment analysis (Vilares et al., 2017), text summarization (Kanapala et al., 2019), code-mixed detection in multilingual documents and on the Web (Bhargava et al., 2016) and machine translation (Sefara et al., 2021). Although language identification has been previously addressed for some of the varieties of Kurdish such as Central Kurdish (Malmasi, 2016), this task is not explored considering all Kurdish varieties.

In addition to the sentences that we extract from our corpus, we collect 3000 sentences for other varieties of Kurdish from the available corpora as follows: Central Kurdish in Perso-Arabic script and Northern Kurdish in Latin script both from the Pewan corpus (Esmaili et al., 2013), Central Kurdish in Latin script from the Wergor corpus (Ahmadi, 2019) and, Zazaki and Gorani sentences from Ahmadi (2020a). Given that types of Northern Kurdish are also written in Perso-Arabic script, particularly in Iraqi Kurdistan, we also collect sentences from online forums and websites that publish in Northern Kurdish written in the Perso-Arabic script. Moreover, we noticed that the script and orthography that is used for Zazaki on its dedicated Wikipedia page[5] is different from the script which is used in Ahmadi (2020a)'s corpus; the latter entirely corresponds to the '*Hawar*' or '*Bedirxan*' system conventionalized for Northern Kurdish (Littell et al., 2016) while the former is influenced by the Turkish Latin script. We did not include the Laki data in this task as the writing system for Laki is yet to be defined in practice. To further diversify the task, we include sentences in Arabic, Persian and Turkish from the Tatoeba datasets as well.[6]

As the baseline system, we evaluate the pre-trained language identifier of fastText (Bojanowski et al., 2017) which can recognize 176 languages including Northern Kurdish, Central Kurdish and Zazaki, respectively with `kmr`, `ckb` and `diq` identifiers. In addition, we train our classifiers where the target classes, i.e. label of the language, include the code of the script, e.g. `ckbarab` and `ckblatn` are used to differentiate between Central Kurdish (`ckb`) text written in the Perso-Arabic and Latin scripts, respectively. Similarly, we consider a classification scenario where the labels are aggregated based on the language code only. As such, we train our model using fastText with the following hyper-parameters: 25 epochs, word vectors of size 64, a minimum and maximum length of char *n*-gram of 2 to 6, a learning rate of 1.0 and hierarchical softmax as the loss function.

Table 5 presents our experimental results of language identification for the selected varieties and scripts. Although the pretrained fastText model–`lid.176` performs poorly, chiefly due to the fact that it has not been trained on our target languages. The results indicate that our trained model per-

---

| Northern Kurdish | Central Kurdish | Southern Kurdish | | Laki | Gorani | Zazaki |
| | | Feyli | Kermanshahi | | | |
| --- | --- | --- | --- | --- | --- | --- |
| *û* (and) | *le* (from, in) | *e* (is) | *û* (and) | *muše* (IND-SAY.PRS-3SG) | *û* (and) | *de* (in) |
| *ku* (that) | *û* (and) | *û* (and) | *we* (and) | *î* (this, these) | *ce* (in) | *û* (and) |
| *li* (from, in) | *bo* (for) | *kî* (that) | *le* (in) | *ye* (a, an) | *be* (to, with) | *ke* (that) |
| *bi* (with, to) | *be* (with, to) | *we* (and) | *abadî* (village) | *arâ* (for) | *be* (that) | *ra* |
| *di* (in) | *ke* (that) | *era* (for) | *naw* (in; name) | *va* (to) | *pey* (for) | *bi* (with) |
| *ji* (from) | *ew* (that) | *ew* (that) | *wegerd* (with) | *mačû* (IND-GO.PRS-3SG) | *y* | *ma* (we) |
| *de* | *Kurdistan* | *kird* (ind-do.pst-3sg) | *ta* (until) | *ya* (this, this one) | *ta* (until) | *xo* (self) |
| *jî* (too) | *Iraq* | *wit* (IND-SAY.PST-3SG) | *ê* (this) | *a* (yes; that) | *î* (this) | *zî* (too) |
| *Kurdistanê* | *em* (this) | *herêm* (region, region of) | *î* (this) | *make* (IND-DO.PRS-3SG) | *Kurdistanî* | *yê* |
| *Iraqê* | *herêmî* (region of) | *Kurdistan* | *weşûn* (after) | *mi* (me, mine) | *her* (each) | *mi* (my) |
| *herêma* (region of) | *serokî* (president of) | *ta* (until) | *bûn* (IND-BE.PST-3PL) | *nâm* (name) | *Turkyay* (Turkey) | *o* (that, it) |

Table 4: The 10 most frequent words in Northern and Central Kurdish and Zaza-Gorani corpora along with our collected data in Southern Kurdish and Laki-Kermanshahi. In addition to frequent function words like prepositions and conjunctions, many words appear related to the topic of the texts, such as *Kurdistan* and *Iraq*.

| Measure | `lid.176` | Our model | | |
| | | language code | language & script code | SDH-unconventional |
| --- | --- | --- | --- | --- |
| Precision | 0.0552 | 0.969 | 0.9638 | 0.25 |
| Recall | 0.0674 | 0.971 | 0.9636 | 0.126 |
| $F_1$ | 0.06 | **0.97** | 0.9634 | 0.168 |

Table 5: Results of language identification with and without the script code (`arab,latn`) included in the label for classification. Unconventional refers to the identification of Southern Krudish text written in Persian script rather than Kurdish. Our models outperform the baseline (pretrained fastText). Measures are computed using the arithmetic mean (also known as macro or unweighted mean).

forms well in both setups where the language code is only provided for the classification task as in `ckb` and also, in the case where the script code is provided as in `ckbarab`. We also evaluate our model on the Southern Kurdish data that is written in the Persian script and orthography prior to being harmonized with the Central Kurdish Perso-Arabic script. An $F_1$ measure of 0.168 reflects the difficulty of this task in a noisy setup as such. A few examples with predictions and heatmaps of the predictions are provided in Table A.1 and Figure A.1.

## 6 Conclusion

Data in general, and corpora in particular, provide a foundation for the preservation and promotion of endangered and under-represented languages in language technology. In this paper, we discuss three approaches for data collection and corpus creation of low-resourced and under-represented languages, namely Southern Kurdish varieties spoken in Kermanshah province (Iran) and Feyli varieties spoken in Iraq. While the Kermanshahi dataset has been collected from a local radio station and by crawling a news website, we collected data for Laki by fieldwork, which despite considerable challenges, seems to be the only solu-tion for a language with near zero online presence. Our approaches can be adopted by other under-represented languages with limited data and without the possibility of fieldwork. We finally provide a brief analysis from quantitative and qualitative perspectives along with the evaluation of language identification for Kurdish and Zaza-Gorani languages with different scripts. Our model can be beneficial to detect texts and collecting more data.

Regarding future work, a data-driven approach can be explored to shed light on the various linguistic differences among the selected languages and varieties. Moreover, as our target languages have been under the threat of linguistic assimilation (Hasanpoor, 1999), particularly through lexical borrowing from Persian, Arabic and Turkish, a new problem transpires which is to determine lexical borrowing. We believe that lexical borrowing detection (Miller et al., 2020) can be further studied in the future thanks to our data. The collected corpora can pave the way for further developments in language technology and cross-lingual studies. Finally, annotating these corpora for other tasks, particularly part-of-speech tagging and named entity recognition can be addressed in the future.

# 7 Limitations

One of the major limitations of the current study is the small size of the collected data with Kermanshahi and Laki having less than 20,000 tokens. Therefore, it is necessary to extend the current data to be able to analyze the languages based on the corpus in a meaningful way (Davies, 2018). The qualitative analysis could be extended to examine the sentence and word length preferences based on the type of text and also, the variety and language.

In order to harmonize the data in Laki and make them comparable with the rest of the Southern Kurdish corpus, transliteration of the Laki corpus is required. However, this requires more discussions among the concerned language community to employ a writing system as the conventional one.

## Acknowledgments

Sina Ahmadi and Antonios Anastasopoulos are generously supported by the National Science Foundation under DEL/DLI award BCS-2109578. The authors are also thankful to the anonymous reviewers.

| Language | Script | Prediction | | Sentence |
|---|---|---|---|---|
| | | `lid.176` | Our's | |
| Northern Kurdish | Bedirxan | KU | KU | Evî jêderî got, ku gundên Reşave, û Kukerê, kevtin ber evê topbarankirinê |
| Northern Kurdish | Central Kurdish | CKB | KU | نزیکی ٦ هەبقایە کێشە د نافیەرا هێزێن سیاسی بێن سەرکەفتی د هەلیژارتن بەردەوامە |
| Central Kurdish | Central Kurdish | CKB | CKB | هەروەها راشیگەیاند، لەتەواوی نەخۆشخانەکاندا برینداران چارەسەریان وەرگرتووە |
| Central Kurdish | Bedirxan | KU | CKB | Parlemanî Turkya dengî be paketî hawdengîy yekêtê Ewrûpa da. |
| Southern Kurdish (Feyli) | Central Kurdish | CKB | SDH | نرخ تەلّای بیگانە و عراقی لە بازارە ناوخۆیەکان ئەرا ئمرۆو دووشەممە داوەزیا |
| Southern Kurdish (Kermanshahi) | Central Kurdish | CKB | SDH | باشد ئاقا نشتبا کردیم، وە خاتر وەزن قافیە شیعرەگە وەتم، بوخشی گەیمان قسەس |
| Southern Kurdish (Kermanshahi) | Persian | FA | FA | امیدواریم لە هر جای استان عزیزمان کرماشان، دەنگمای شنوین، دلخوەش بیون |
| Zazaki | Bedirxan | KU | ZZA | Şima seba îadeyê heqanê şarê Dêrsimî û qedînayîşê polîtîkayanê teda |
| Zazaki | Wikipedia | DIQ | ZZA | Agariyaki yew zıwanê Hındıstaniyo. Aidê gruba Zıwanê Mundayo. |
| Gorani | Central Kurdish | CKB | HAC | ئازاڵ و سەربەوێژ و شابان و نمونە بۆ و هەریاسە داراو پلە وپایەی بەرزی کۆمەلّایەتی |

Table A.1: A few examples in the selected languages along with the predictions of fastText's pretrained models (`lid.176`) in comparison to those of our model trained using fastText on our collected data. Northern Kurdish (KU), Central Kurdish (CKB) and Southern Kurdish (SDH) are used along with Gorani (HAC) and Zazaki (ZZA) in various scripts. DIQ refers to the script that is used for Zazaki on Wikipedia.

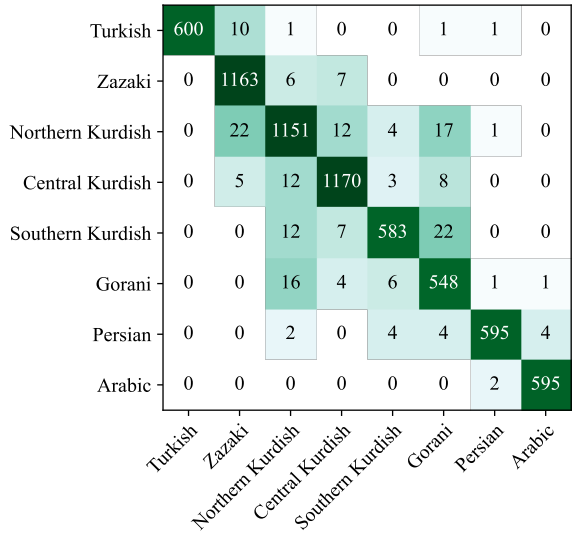

(a) Classification with language codes

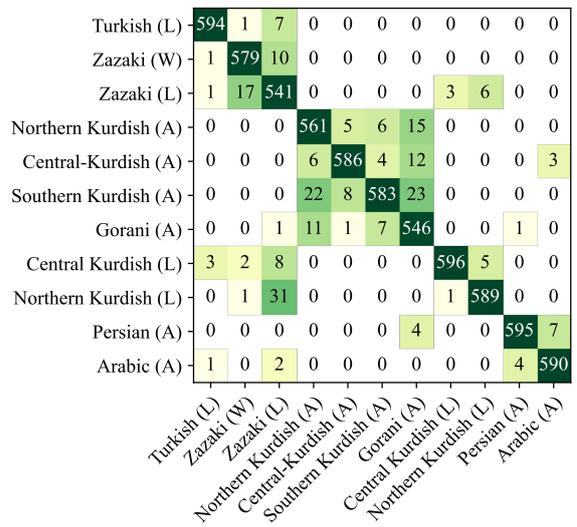

(b) Classification with language and script codes

Figure A.1: Language identification of Kurdish varieties and Zaza-Gorani when considering the script as a label (to the right) and without the script (to the left). Script codes are shown as L, A and W for Latin, Arabic and Zazaki Wikipedia. The number of classifications is annotated. Horizontal and vertical axes refer to reference labels and model predictions, respectively.